# Predicting Survival of Hemodialysis Patients using Federated Learning

A Nation-wide study on a subsample from NephroPlus, India's largest network of dialysis centers.

**Abhiram Raju**
*Chirec International School*
Hyderabad, India
abhiramraju7@gmail.com

**Praneeth Vepakomma**
*MBZUAI, ADIA Lab and IDSS, MIT*
Mohamed bin Zayed University of Artificial Intelligence and Massachusets Institute of Technology
Cambridge, U.S.A and Abu Dhabi, U.A.E
Vepakom@mit.edu

*Abstract*—Hemodialysis patients who are on donor lists for kidney transplant may get misidentified, delaying their wait time. Thus, predicting their survival time is crucial for optimizing waiting lists and personalizing treatment plans. Predicting survival times for patients often requires large quantities of high-quality but sensitive data. This data is siloed and since individual datasets are smaller and less diverse, locally trained survival models don't perform as well as centralized ones. Hence, we propose the use of Federated Learning (FL) in the context of predicting survival for hemodialysis patients. FL can have comparatively better performances than local models while not sharing data between centers. However, despite the increased use of such technologies, the application of FL in survival and even more, dialysis patients remains sparse. This paper studies the performance of FL for data of hemodialysis patients from NephroPlus, India's largest network of dialysis centers.

## I. INTRODUCTION

India has one of the highest numbers of patients receiving chronic dialysis worldwide: about 175,000 patients as of 2018 [1]. This largely consists of hemodialysis (HD) patients and with an additional 3.4 crore dialysis in India every year [2], there is a growing burden on facilities and families. Given the large number of dialysis patients, the number of those requiring transplants is also rising, creating longer waitlist times. Thus, there is a large need for being able to predict survival times of these patients. To do this, several studies [3-6] have been carried out using survival (time-to-event) analysis to predict the time to an observed event occurring for dialysis patients using a variety of statistical and machine learning techniques. Although many of them do focus on survival within local context like [6], very few do so in an Indian context. For the Indian demographic, a national benchmark for dialysis patients has only recently been set by [7].

Such studies come under survival or time-to-event analysis and have found considerable success in the hemodialysis space. However, in real-world situations, data is often distributed or may not be large enough in size to train a model that generalizes well to real-world scenarios. Given privacy constraints, it's often difficult to share data between different dialysis centers and even hospitals. Hence, we propose the use of Federated Learning [8][9]. Federated Learning (FL) offers a method for centers to collaborate with one another without sharing data directly. This takes place by training multiple local models, each trained on a client's private data.

Federated Learning has been applied in Survival Analysis primarily upon Cox Proportional hazards models [10-12]. However, in recent years there have been studies carried out which apply FL for a variety of machine learning and deep learning applications in survival and healthcare [13-15].

However, most of the studies carried out don't include real-world data from multiple clients or centers. Furthermore, FL remains relatively sparse in the field of dialysis. As per our best knowledge, this is the first study that applies federated learning on a large real-life dataset for hemodialysis patients. This would also be the first to leverage FL for healthcare in an Indian context. We did this through data from NephroPlus, India's largest private dialysis network [16]. Across this network, information was collected from 2013-2020 in studies to record electronic medical records (eMR), capturing valuable demographic and clinical data, vital for predicting the survival of the dialysis patients.

## II. BACKGROUND

Survival analysis or time-to-event analysis is a branch of statistics whose primary objective is to model and analyze the time until an event of interest occurs. The event of interest is often referred to as a failure or death and is our focus.

*A. Survival Function*

The survival function, *S(t)*, represents the probability that an individual survives beyond time *t*:

$$S(t) = P(T > t) \quad (1)$$

where T is a random variable denoting the time of the event and the function represents the probability that an individual has not experienced an event up to time t.

In our context, survival analysis is vital to achieve the goal of predicting the survival time of a new hemodialysis patient given their associated covariates. The survival function is estimated from our survival dataset which consists of 3 key variables: $x_i$, $t_i$ and $\delta_i$ for the *i*-th sample. For this patient, $x_i$ represents the feature vector while $t_i$ is the minimum between the censoring time and the actual event occurrence time. $\delta_i$ is the event indicator, where $\delta_i = 1$ signifies that the event occurred for the subject at time $t_i$, and $\delta_i = 0$ indicates that the event did not occur, meaning the record is censored.

*B. Federated Learning*

Federated Learning (FL) refers to a Machine Learning Scenario where different clients collaborate using their own data through a central server. FL provides a method for clients to coordinate with one another despite never being shared. This can considerably improve privacy as models can simply share their weights, helping avoid problems caused by sharing sensitive data.

One of the first ones to be used is Federated Averaging (FedAvg) [8]. Here, the model weights are averaged based on client dataset size to prioritize losses with low variance that are evaluated on larger datasets. FedAvg uses 2 steps: a broadcast step and an aggregation step. In the broadcast step, the server selects the clients and sends its current global model weights, *w*. In the aggregation iterations, once each client has set its weights as the global model weights, each model is trained on its own local data to minimise a local loss function. Once the model has done this for a sufficient number of epochs, all the model's weights are averaged using FedAvg to compute the update and set the global model's weights. The weighted average is calculated as follows:

$$w^{(t+1)} = \sum_{k=1}^{K} \frac{n_k}{N} w_k^{(t+1)} \quad (2)$$

where $n_k$ is the number of data samples on the *k*-th client, and $N \sum_{k=1}^{K} n_k$ is the total number of data samples across all clients. We wanted to evaluate the effectiveness of a Federated Learning framework to achieve this goal of predicting survival times of HD patients through a decentralized approach. If the performance is comparable, given FL's advantages of improved privacy and collaboration, it may suggest that many dialysis centers and zones could be better off with Federated Learning.

## III. METHODOLOGY

*A. Nationwide Client Data*

The study analyses data collected by Nephroplus, India's largest dialysis network. We are the among the first to analyse hemodialysis data on a large scale and will be focusing on data collected from 244 centers in India from 2013 to 2020. Out of the 183063 patients here, 24052 patients fulfilled our criteria to be included in the study. Of these 24052 patients, 14805 patients are right-censored while 9247 patients have expired.

We have included 17 different variables that include a range of demographic and medical data. The features are: Guest Type, Age, Gender, End-Stage Renal Disease, Hematuria, Alcohol Intake, Diabetes, Dislipidemia, ECG Abnormality, Heart Attack history, Heart failure history, Exercise levels, Peripheral Vascular Disease history, Stroke history, Hypertension history, Smoking Status and Visual Analog Scale (VAS) value.

Out of the large number of centers patients were registered at, each was under one of six zones based on geographical location and data samples. Each zone: North, South, East, West, Andhra Pradesh and Bihar was considered as a client with its own unique dataset from the centers it oversees. This approach simulated a situation where there is a regional authority overseeing each set of centers and a global authority whose federated learning framework these clients are joining. The number of patients registered at that zone as well as the number of censored/expired patients is shown in Table I.

For predicting the survival times of HD patients in our study, we have 4 models: Cox Proportional Hazards model, DeepSurv, Cox-nnet and Random Survival Forests.

Table I: Patient Data and Observation Type by Client Zone

| Client(Zone) | No. of Total Patients | No. of Expired Patients | No. of Censored Patients |
|---|---|---|---|
| North | 5094 | 2017 | 3077 |
| South | 5046 | 2149 | 2897 |
| East | 3494 | 1152 | 2342 |
| West | 3085 | 1202 | 1883 |
| Andhra Pradesh | 6032 | 2234 | 3798 |
| Bihar | 1301 | 493 | 808 |

*B. Models Used*

Semi-parametric models model the instantaneous risk or hazard of experiencing the event at each given momentary time period. They predict a hazard function, $\lambda(t)$, given that the individual has survived up to time *t*:

$$\lambda(t) = \lim_{\Delta t \to 0} \frac{P(t \leq T < t+\Delta t | T \geq t)}{\Delta t} = \frac{f(t)}{S(t)} \quad (3)$$

The Cox Proportional Hazards model is one of the most used statistical, non-parametric methods for modelling an individual's survival given their set of features. As described in [17], it estimates the effect of each covariate on the survival. It does so by semi-parameterizing a hazard function $\lambda(t)$ as:

$$\lambda(t|x) = \lambda_0(t) \cdot e^{g(x)} \quad (4)$$

Here, *t* is the time period while *x* represents each covariate and g(x) is the risk function. The baseline hazard, $\lambda_0(t)$ has been parametrized by the cox-proportional hazards model, enabling it to be flexible. This $\lambda_0(t)$ can then be evaluated using the Nelson-Aalen Estimator [18] and in the logarithmic risk function, $e^{g(x)}$, the linear risk function g(x) is modelled as $g(x)=\beta^T x$.

One of the key features of the Cox model is its use of the partial likelihood, which allows for efficient estimation of the regression coefficients, β under the assumption that the baseline hazard function can be left unspecified. Given *n* individuals, $t_i$ is the observed time for the *i*-th individual, and $\delta_i$ is an event indicator variable. Using Breslow's estimate [19] for handling tied event times, the partial likelihood function for the Cox model is given by:

$$L(\beta)= \prod_{i=1}^{n} \left( \frac{e^{g(x_i)}}{\sum_{j \in R(t_i)} e^{g(x_j)}} \right)^{\delta_i} \quad (5)$$

where $R(t_i)$ is the risk set of individuals who are still at risk of experiencing the event just before time $t_i$. To do this, the negative partial log-likelihood can be modelled as a log function:

$$loss=- \sum_{i=1}^{n} \left[ g(x_i) - \log \left( \sum_{j \in R(t_i)} e^{g(x_j)} \right) \right] \quad (6)$$

Thus, the model tries to maximise the partial likelihood function and minimise the negative log loss. However, the assumption here is that the hazard ratio for a given covariate remains constant over time. Hence, several non-linear machine learning and deep learning models have been utilized in this context.

In DeepSurv[20], the risk function from (4) is now $\lambda(x) = \phi_w(x)$ where $\phi_w$ is a single-output neural network that extends the CoxPH model and outputs the hazard in a non-linear manner. Extending the Cox Proportional Hazards regressor, DeepSurv too is semi-parametric and uses the log loss function to optimise the partial likelihood while the baseline hazard function may remain undefined.

To compare another similar neural network model, we have also evaluated a modified version cox-nnet that is suited for electronic health record(EHR) data [21]. Cox-nnet typically emphasizes architectural features such as the inclusion of batch normalization and dropout layers, aiming to enhance model generalization and prevent overfitting.

Random Survival Forests (RSF), developed by [22], is an ensemble learning method used for survival analysis, particularly well-suited for handling high-dimensional data and complex interactions among covariates. RSF extends the concept of decision trees to analyze time-to-event data. Each tree in the forest is grown using a bootstrap sample of the data, and the final survival prediction is obtained by averaging the survival estimates from all trees. RSF is non-parametric and does not assume proportional hazards, making it flexible in capturing non-linear relationships.

For our analysis we have chosen these 4 models and compared their performance in local client-level vs federated context. Our four model types are shown in table II.

Table II: Comparison of Models Based on Linearity and Proportional Hazards

| Models | Linear | Proportional Hazards |
|---|---|---|
| CoxPH | Yes | Yes |
| DeepSurv | No | Yes |
| Cox-nnet | No | Yes |
| RSF | No | No |

*C. Federating the Models*

To investigate the effect of incorporating federated learning techniques for predicting the survival of hemodialysis patients, each of our local models has been applied in a federated context.

*a) Federating CoxPH:* Most applications of FL in Survival Analysis focus on optimizing the FL it in the context of the Cox Proportional Hazards model. After fitting the local CoxPH models, we used Naive Global Models Parameters Averaging [23] approach by averaging each of the *β* coefficients for every variable from the local CoxPH models using the algorithm:

$$\beta_{global}= \frac{\sum_{k=1}^{K} r_k \beta_{local,k}}{\sum_{k=1}^{K} r_k} \quad (7)$$

where $\beta_{global}$ is the Global beta coefficient for a given feature and $\beta_{local,k}$ are Local beta coefficients from the *k*-th center of a total *K* centers. $r_k$ is the weight factor for the *k*-th center to show the size or importance of the data from that center. Once the $\beta_{global}$ have been computed, broadcast step is redone to set the coefficients for all the local CoxPH models.

*b) Federating Neural Networks- DeepSurv and Cox-nnet:* To train the local DeepSurv and Cox-nnet models for each client, we fitted each for a 100 epochs and carried out a grid search to find the optimal hyperparameters. Each of the models was trained to minimize the cox partial likelihood log-loss (6) while predicting the hazard for patients.

For the federated versions of DeepSurv and Cox-nnet, we performed a FedAvg algorithm where the weights were aggregated using the weighted federated averaging described in equation (1). The global model was then set using these aggregated weights and the broadcasting step was reiterated.

*c) Federating Random Survival Forests*

The local Random Survival Forest models were also trained for each client to predict the survival times of the dialysis patients. Once they were trained, we adopted methods from [24] and used importance-based sampling to create this. Each client *k* maintains an ensemble of survival trees denoted as $M_k=\{T_1,\ldots,T_{N_k}\}$ where $N_k$ represents the number of trees within the client's ensemble.

Each client's decision trees were sorted based on performance. A central server determined the number of trees it required from each client based on the importance assigned to each of them. The top-performing trees were taken from each client's forest until a tree counter is no longer less than or equal to a required number of trees. Using these model's estimators, the global random survival forest was set with:

$$M_s = \bigcup_{k=1}^{K} M'_k \qquad (8)$$

where $M'_k$ represents the model from the *k*-th client and $M_s$ is the aggregated global model.

Overall, these methods of applying federated learning across NephroPlus's different client zones is shown in Fig. 1.

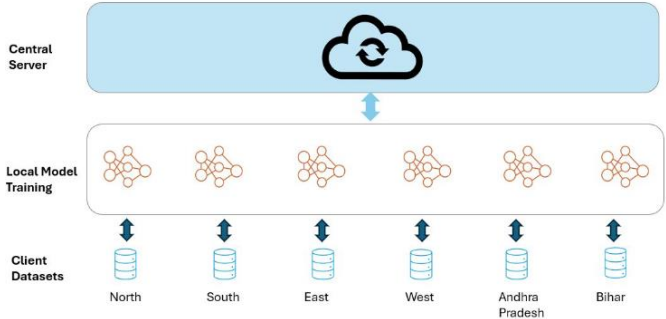

Figure 1: Federated Learning Setup

## IV. RESULTS

Each of these global federated model types (Federated Survival Forests, Federated DeepSurv etc.) was then evaluated for local client data to compare the performance of federated learning models against typical local model performance to determine whether or not it is beneficial to join the FL framework. The testing data was obtained locally for each client through an 80/20 train/test split.

We wanted to explore the effect of FL in the context of predicting hemodialysis survival times. These models were each evaluated using Harrell's concordance index score on a scale of 0 to 1. The results of our runs can be seen in Table III. As depicted, the bold values refer to the instances where the Federated Learning model performs better than its local counterpart. Each client has at least 1 Federated Learning model outperforming its corresponding local model. North, South, AP and West clients have a greater than or equal performance in 2/4 models meanwhile East and Bihar do so in 1/4 models.

Table III: Performance Metrics of models across clients

| Models | North Local | North Federated | South Local | South Federated | East Local | East Federated | West Local | West Federated | AP Local | AP Federated | Bihar Local | Bihar Federated |
|---|---|---|---|---|---|---|---|---|---|---|---|---|
| CoxPH/FedCoxPH | 0.587 | **0.587** | 0.573 | **0.579** | 0.647 | 0.639 | 0.595 | 0.594 | 0.625 | 0.613 | 0.607 | 0.488 |
| DeepSurv/FedDeepsurv | 0.634 | 0.594 | 0.607 | 0.568 | 0.647 | 0.567 | 0.592 | 0.538 | 0.642 | 0.576 | 0.564 | **0.568** |
| Cox-nnet/FedCoxnnet | 0.604 | 0.576 | 0.570 | **0.573** | 0.576 | 0.544 | 0.547 | **0.549** | 0.570 | **0.570** | 0.579 | 0.544 |
| RSF/FedSurf | 0.621 | **0.639** | 0.612 | 0.611 | 0.671 | **0.683** | 0.586 | **0.618** | 0.655 | **0.655** | 0.591 | 0.568 |

The local and federated learning models with the best performance were RSF and Federated RSF due to their ability to handle complex non-linear relationships, assign feature importances and avoid overfitting. Hence, since clients are likely to opt for the best performing local and federated models, we have displayed the performance of RSF against FedSurf in Fig. 2.

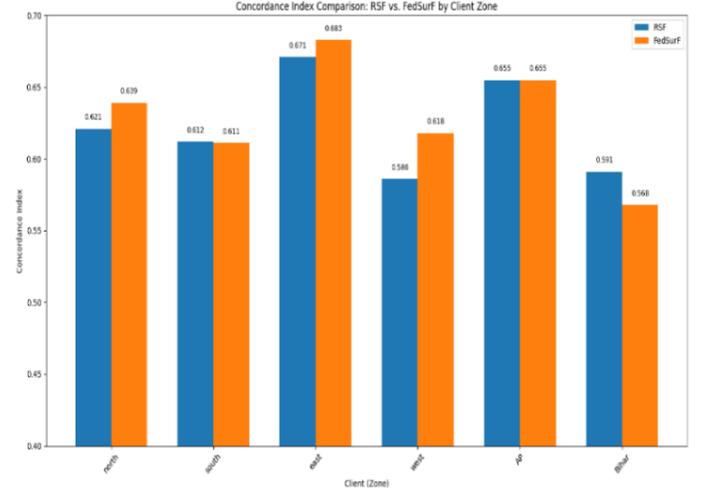

Figure 2: Bar Charts depicting Performance of federated survival forests against random survival forests

Furthermore, on a general context across all models run, it was a Federated Learning model that had the best performance for 4 out of 6 clients: FedSurf had the best performance of all models in the North, East, Andhra Pradesh and West zones. On the other hand, local RSF and local CoxPH models had the best performance for South and Bihar zones respectively. This trend, shown in table IV, demonstrates that 4 out of 6 clients have an increased incentive to use a federated learning model to maximise performance.

Table IV: Best Performing model type for each client

| Client | Best Performing Model Type |
|---|---|
| North Zone | Federated |
| South Zone | Local |
| East Zone | Federated |
| West Zone | Federated |
| Andhra Pradesh Zone | Federated & Local |
| Bihar Zone | Local |

This indicates that Hemodialysis regional clients have an incentive to use a federated learning model not only for a more secure and robust model but also for improved accuracies. The charts in fig. 2 demonstrate how for some clients, local and federated models may have similar concordance index scores. However, clients may find a relatively small accuracy trade-off for some models to be compensated by the other benefits of FL. By following a federated learning model, these clients not only have increased privacy but are also exposed to more heterogenous data for a more diverse dataset. Since they are likely to use the best performing models, table IV shows how most of the top performing models overall are federated learning ones.

## V. Conclusion and Improvements

The aim of the study was to apply federated learning to the dialysis space and compare federated and local models in predicting the mortality of hemodialysis patients. Our results showed that federated learning has comparable performance to models trained on local data and in many cases, outperforms it as well. The results displayed that Random Survival Forests as well as neural network models can improve their performance through FL. Furthermore, we found that the feature importances generated by RSF models differed from client-to-client, depicting how different regions may be biased towards certain patterns but FL models can generalize well using varied models from different zones. Finally, although some models like DeepSurv didn't perform as well in a federated context, it was likely due to the fact that the data was very heterogeneous with different clients contributing different proportions to the global model.

Our results seem positive regarding federated learning for survival analysis of Hemodialysis patients and suggest there is scope to explore similar experiments on an even larger scale. This can potentially increase the improvements when using FL. Carefully expanding the number of features to include more clinical and demographic data could also result in more accurate results. Furthermore, with local datasets of more similar sizes, clients like Bihar with fewer numbers of patients would also have greater performance. Finally, FL is a great step towards privacy as unlike distributed computing and other networking applications, medical centers wouldn't have to send data. To further improve privacy, differential-privacy is a great option for future studies.

## Acknowledgment

We would like to thank NephroPlus for providing the data in this study and also to the patients who consented for their data to be used for research purposes. We also thank MBZUAI and ADIA lab for the support in this research.